\documentclass[a4paper,final,10pt]{article}
\usepackage[english]{babel}			 
\usepackage[dvips]{graphicx}		
\usepackage{amsmath}   
\usepackage{amsfonts}
\usepackage{textcomp}
\usepackage{nicefrac}

\usepackage{cite}
\newcommand{\be}{\begin{equation}}
\newcommand{\ee}{\end{equation}}
\newcommand{\bg}{\begin{gather}}
\newcommand{\eg}{\end{gather}}

\newcommand{\bq}{\begin{quote}}
\newcommand{\eq}{\end{quote}}

\begin{document}
\title{\bf Magneto-mechanical actuation model for fin-based locomotion}    
\author{Juan Pablo Carbajal and Naveen Kuppuswamy}
\thanks{Artificial Intelligence Laboratory.\\ University of Z\"{u}rich.\\ e-mail: \{carbajal,naveenoid\}@ifi.uzh.ch}              

\date{\today}                           

\maketitle                              
\setlength{\topmargin}{0cm}
\abstract{In this paper, we report the results from the analysis of a numerical model used for the design of a magnetic linear actuator with applications to fin-based locomotion. Most of the current robotic fish generate bending motion using rotary motors which implies at least one mechanical conversion of the motion. We seek a solution that directly bends the fin and, at the same time, is able to exploit the magneto-mechanical properties of the fin material. This strong fin-actuator coupling blends the actuator and the body of the robot, allowing cross optimization of the system's elements.

We study a simplified model of an elastic element, a spring-mass system representing a flexible fin, subjected to nonlinear forcing, emulating magnetic interaction. The dynamics of the system is studied under unforced and periodic forcing conditions. The analysis is focused on the limit cycles present in the system, which allows the periodic bending of the fin and the generation of thrust. The frequency, maximum amplitude and center of the periodic orbits (offset of the bending) depend directly on the stiffness of the fin and the intensity of the forcing; we use this dependency to sketch a simple parameter controller. Although the model is strongly simplified, it provides means to estimate first values of the parameters for this kind of actuator and it is useful to evaluate the feasibility of minimal actuation control of such systems.}

\section{Introduction}
In the last two decades underwater fin-based propulsion has been a topic of intense research. Theoretical models and simulations based on a multiplicity of numerical methods have been developed and reported \cite{Cheng98, McMillen06,  Eldredge09, KANSO09, Kanso09-2, Shukla07, ELDREDGE08, Alben10} as well as experiments performed on artificial platforms and with animals \cite{Ahlborn97, Lauder07-1, Epps09, Deng05}. Theoretical studies such as\cite{Cheng98, McMillen06} focus on models of the viscoelastic body, while\cite{Eldredge09, KANSO09, Kanso09-2} treat the body-fluid coupling and the emergence of locomotion. In \cite{Shukla07, ELDREDGE08,Alben09} the main interest is to understand how passive thrust is generated in vortex wakes, a question that remains open. Despite the considerable work done, experimental results are not fully understood and several numerical simulations are yet to be validated. Influenced by the activity on the field, we address a parallel problem related to robotics, \emph{actuation}. 

Fins as a tool for locomotion offer several appealing properties with respect to propellers. From an environmental point of view fins reduce sound pollution characteristic of propellers\cite{Richard95}. Additionally, a flexible body offers the possibility of extracting energy from the environment, as shown in the technological study presented in \cite{Allen01}. This idea is related to the fact that trouts exploit vortex wakes (shed by obstacles in the flow) to reduce the cost of swimming\cite{Beal06} (see \cite{Liao07} for a review). These aspects are of primary relevance in situations where low environmental impact and mimicry are important, as in pipes maintenance routine, or for underwater life observation (scientific naiveness may make us forget about military applications, we provide this short caution.). Moreover, in environments where moving parts may be clogged up due to fouling, rotatory propellers may be unfeasible for locomotion.

In the field of bio-inspired robotics, actuation for swimming robots using fin-like propulsion is usually implemented by the use of rotatory electric motors to control the fin swing angle. However, The presence of motors hinders miniaturization and integration of actuators into the robot structure, a requirement for flexible machines with deformable body. At the same time this imposes a mechanical conversion of rotations into oscillatory linear motion, the complexity of which becomes an issue at smaller size. Moreover, minimizing or eliminating the number of moving parts required for actuation simplifies maintenance to a large extent. To tackle the mentioned difficulties we are directing our research towards new ways of actuation. Herein we report a simple mathematical model and the numerical analysis of a possible alternative. 

\begin{figure}[htbp]
\label{fig:Setup}
\includegraphics[width=.8\textwidth]{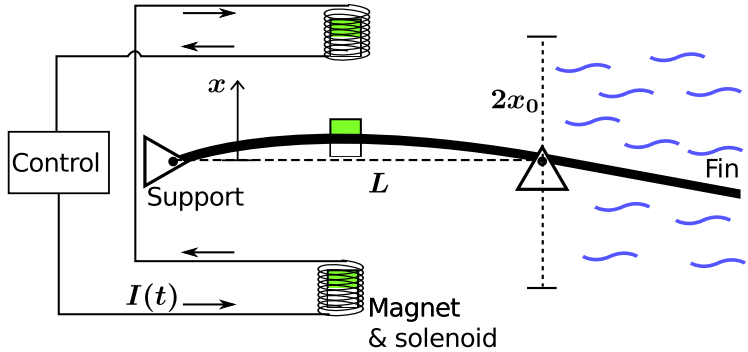}
\caption{Schematic of the system described by equations (\ref{eq:dynamics}). A beam is used to model the fin and it is simply supported at two points. The actuation is done by means of a combination of permanent magnets and solenoids. The distance between the supports defines the rigidity of the actuated section.}
\end{figure}

We start by considering a robot composed of a hull and a fin attached to it. The fin is modeled as an elastic beam (see \cite{McMillen06} or \cite{Cheng98} for more detailed models), which we want to set into oscillatory motion. In the setup shown in Fig. \ref{fig:Setup}, we choose to support the beam at two points. The first support is at one edge of the beam and stands for the hull of the robot. The second support is placed at some intermediate point in the beam. The section of the beam beyond this second support is meant to generate the thrust by interacting with the surrounding fluid. The actuation is done in the section of the beam between the two supports by means of a combination of permanent magnets (one of them attached to the beam) and solenoids. In the configuration chosen, the permanent magnets serve to increase the compliance of the system and to reduce the force that needs to be actively applied by the solenoid. The distance between the supports defines the rigidity of the actuated section and could be tuned for optimal energy transfer. Similar working principles are described in patents of electric razors, and of active dampers of oscillations for digital cameras lenses (in these contexts the actuator is often called  {\it motor} or {\it electromagnetic spring}). Similarly, the control of the resonant modes of a structure is a commonplace problem in structural dynamics\cite{Vakakis09}. It is noteworthy that all these techniques exploit (to be more efficient) or require (to be implemented) knowledge of the resonant modes of the system under study. 
\section{Dynamic Model}
\label{sec:dynamic_model}
The displacement of the magnet in the fin (we will refer to this point as the {\it fin magnet}), can be modeled by a spring-mass system under the effect of an external force field. Considering only one dimensional motion, the system is written as,
\begin{equation}
\begin{split}
\dot{x} &= v \\ 
\dot{v} &= \frac{F_2(x) + F_1(x) + F_{s1}(x)+ F_{s2}(x)}{m} - \frac{\mathcal{K}}{m} x - \frac{\Gamma}{m} v,
\end{split}
\label{eq:dynamics}
\end{equation}
\noindent where $x$ is the displacement of the fin magnet, $m$ is an effective moving mass, $\mathcal{K}$ represents an effective elastic constant of the fin setup and $\Gamma$ is used to include dissipation. The $F_i$ and $F_{si}$ terms are forces acting on the fin due to the external magnets and the solenoids, respectively.

Magnetic forces can be highly complex; to keep our model as simple as possible we approximate each magnet as a point magnetic dipole, which is a good approximation when the distances are significantly bigger than the size of the magnet in the direction of the magnetization\cite{Vokoun09}. In this situation the force can be expressed as follows,
\begin{equation}
F_i(x) = - \frac{\mathcal{C}_i}{\left(x-x_i\right)^\alpha} \operatorname{sign}(x-x_i),
\label{eq:Magforce}
\end{equation}
\noindent where $\mathcal{C}_i$ is a constant that depends on the magnetic moments of the magnets and their geometry, positive values represent attractive forces and negative values repulsive forces. In the case of the solenoids, this constant depend also on the current, i.e. $\mathcal{C}_{si}(I)=C_{si}I(t)$ (the index $s$ refers to solenoid). The position of the external magnet (or solenoid) measured from the rest position of fin magnet is $x_i$. Henceforth we define $k=\nicefrac{\mathcal{K}}{m}$, $\gamma=\nicefrac{\Gamma}{m}$, $c_i=\nicefrac{\mathcal{C}_i}{m}$ and $c_{si}=\nicefrac{C_{si}}{m}$. Additionally, we assume that the deflected fin does not reach the external magnets, in mathematical terms this is expressed as $x \in (x_1,x_2)$. 

Next we study the dynamics of the system without actuation, $I(t) \equiv 0$. The expression for the fixed points $x^*$ is obtained by equating system (\ref{eq:dynamics}) to zero. The second equation yields
\begin{equation}
c_2\left(x^*-x_1\right)^\alpha - c_1\left(x^*-x_2\right)^\alpha - k x^*\left[\left(x^*-x_1\right)\left(x^*-x_2\right)\right]^\alpha = 0,
\label{eq:fppoly2}
\end{equation}
\noindent where the assumption $x \in (x_1,x_2)$ was used to determine the signs.

\paragraph*{Linear Stability Analysis.}
To classify the fixed points, we calculate the trace and determinant of the $2 \times 2$ Jacobian matrix $J$ of (\ref{eq:dynamics}). These are given by 
\begin{eqnarray}
\label{eq:Trace}\operatorname{Tr}(J) &=& -\gamma \\
\label{eq:Det}\operatorname{Det}(J) &=& k + \alpha \left[ \frac{c_2}{\left(x - x_2\right)^{\alpha + 1}} - \frac{c_1}{\left(x - x_1\right)^{\alpha + 1}}\right].
\end{eqnarray}

In general the fixed points of the system will be saddle-nodes, centers or spirals, depending on the value of the parameters $\gamma$, $k$, $c_i$ and $x_i$. However, the position of the fixed point (i.e. the solutions of (\ref{eq:fppoly2})) are independent of $\gamma$. 

To proceed with the analysis we introduce further assumptions. The exponent $\alpha$ depends on the arrangement of magnets\cite{Vokoun09}. Here we will consider identical cylindrical magnets placed symmetrically with respect to the rest position of the fin magnet and with dipoles parallel to it (attracting); hence $\alpha=4$, $c_1=c_2=c > 0$ and $x_2=-x_1=x_0 > 0$. By neglecting dissipation, i.e. $\gamma = 0$, we set the trace of the Jacobian to zero. Consequently, the fixed points are either saddle-nodes or centers, depending on the sign of (\ref{eq:Det}).  Using these assumptions to simplify the equality (\ref{eq:fppoly2}) we obtain,
\begin{equation}
x^*\left[8cx_0\left(x^{*2}+x^2_0\right) - k \left(x^{*2}-x^2_0\right)^4\right] = 0,
\label{eq:fppoly3}
\end{equation}
\noindent rendering evident that $x^*=0$ is one of the fixed points, in consequence of the symmetry of the problem. The determinant (\ref{eq:Det}) at this point is,
\begin{equation}
\operatorname{Det}(J)\vert_{x^*=0} = k - 8\frac{c}{x_0^5},
\label{eq:tradeoff}  
\end{equation}
\noindent which is positive for $\nicefrac{c}{k} < \nicefrac{x_0^5}{8}$, and the origin is a center. Although any real system will not show centers without actuation (due to dissipation) their position will match the pole of the spirals observed.

\begin{figure}[htbp]
	\includegraphics[width=.7\textwidth]{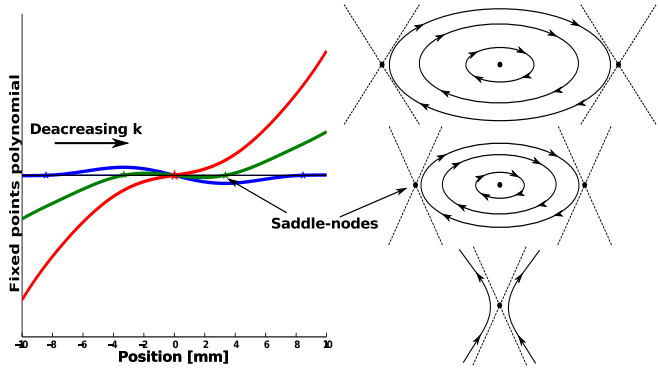}
	\caption{Plot of the polynomial defined in (\ref{eq:fppoly3}) for different values of $\nicefrac{c}{k}$. The star symbols mark the position of the fixed points. The phase portraits to the right show that the saddle-nodes define a limit for the amplitude of the orbits. The figure illustrates the trade-off between the rigidity of the fin and the intensity of the magnetic interaction.}
	\label{fig:fppoly}
\end{figure}

It can be shown that the nonzero solutions of (\ref{eq:fppoly3}) are saddle-nodes. Displacements beyond the saddle-nodes will bring the fin magnet into a region where the attraction is stronger than the elastic restitution, causing the fin to stick to the closest magnet. The saddles establish a natural limit for the maximal amplitude of the orbits of the system. To illustrate these ideas, we show in Fig. \ref{fig:fppoly} three plots of the polynomial defined by (\ref{eq:fppoly3}) for different values of the ratio $\nicefrac{c}{k}$, together with phase portraits of the system. The figure depicts the trade-off between the rigidity of the fin and the interaction of the fin magnet and the permanent magnets. Keeping the $x_0$ fixed, the stronger the magnets (or the more compliant the fin), the smaller the region where the system can present stable orbits. At the critical ratio $\nicefrac{c}{k} = \nicefrac{x_0^5}{8}$, the saddle-nodes collide at the origin and the center is transformed into a saddle-node.
\section{Actuation and Control}
\label{sec:actuation_and_controller}
As mentioned before, dissipation will reduce the amplitude of the oscillations. Therefore, to keep the system close to the desired trajectory we need to pump energy into it. To do this, we have placed a solenoid surrounding both magnets such that they can increase or decrease the interaction with the fin magnet. Both solenoids are constructed similarly but arranged anti-parallel to each other ($c_{s2}= - c_{s1}=c_s > 0$). We place them as close as possible to the fin magnet, for example near the saddle-nodes of the system. In order to determine the parameters of the solenoid needed to drive and control the system, we use a simple PID controller which can regulate the applied force. To this end, we rearrange the terms of (\ref{eq:dynamics}) and write them as,
\begin{equation}
\begin{split}
&\dot{v} = \frac{c}{\left(x-x_0\right)^4} - \frac{c}{\left(x+x_0\right)^4} + F_c(t) - k x - \gamma v\\
& F_c(t) = k_p e(t) + k_d\dot{e}(t) + k_i \int_0^t e(s) ds\\
& e(t) = x_d(t) - x(t).
\end{split}
\end{equation}
\noindent Where $x_d(t)$ is the desired displacement. The effort required to drive the system into steady oscillations, depends on the appropriate choice of $k_p$, $k_d$, $k_i$.
\section{Parameter values}
In the following sections we define the values of the parameters used for the numerical simulation of the platform. We give a brief description on the assumptions and criteria used to select them. In Table \ref{tab:constants} we summarize the information. 
\paragraph*{Magnets.}
The constant for the force (\ref{eq:Magforce}) in the case of cylindrical magnets magnetized along their length $\ell$ is,
\begin{equation}
\mathcal{C} = \frac{3\mu_0}{2\pi}\left(\frac{B_r}{\mu_0}\right)^2\left(\pi R^2 \ell\right)^2,
\label{eq:Magconstant}
\end{equation}
\noindent where $\mu_0$ is the permeability of vacuum, $R$ is the radius of the magnet and $B_r$ is the remanence of the magnet (a value available from manufacturers). The factors in parenthesis represent the magnetization and the volume of the magnet, they are squared because we are assuming both interacting magnets are equal. Using values from commercially available Neodymium magnets we have calculated $\mathcal{C} = 2.460 \times 10^{-10} N\cdot m^4$.
\paragraph*{Elastic constant.}
As can be seen in Fig. \ref{fig:Setup} we model the fin using a beam simply supported in two points. The supports are  separated by a distance $L$. The fin magnet is placed at the point $y$ and the elastic stiffness there can be written as,
\begin{equation}
\mathcal{K} = EJ \frac{3L}{y^2(L-y)^2},
\label{eq:stiffness}
\end{equation}
\noindent where $E$ is the Young's modulus of the material and $J$ is the second moment of area of the beam. As discussed before, the behavior of the system depends on the relation between the elastic constant of the fin and the strength of the magnets. The current setup allows tuning the elasticity of the fin by setting different materials and profiles of the fin, or by moving the fin magnet, or by changing the distance between the supports. If needed, the setup could be transformed into a cantilever by removing the second support. To provide good ranges of elasticity we use Low-density polyethylene plastic with $E \approx 0.2 \times 10^9 Pa$ and $L = 30 mm$. The fin has width and thickness of $10 mm\times 0.5 mm$, respectively. The fin magnet is placed in the middle of the two supports. This values yield $\mathcal{K} = 37.03 N\cdot m^{-1}$.
\paragraph*{Damping and mass.}
When a body moves in a liquid, it transfers kinetic energy to the surrounding fluid reducing the acceleration it presents corresponds to the one observed on a body with higher mass. This phenomena, known as {\it added mass}, can be estimated using models as the one presented in\cite{Yadykin03}. However, we postpone a detailed description for future work and simply consider a total mass 300 times bigger than the mass of the fin and the magnet together, $m = 84.3 g$. 
Estimation of the damping $\Gamma$ without an experimental setup is not straightforward, therefore we use damping ratios in the range $Q \in [0, 0.5]$. Where $Q=0$ means no damping and a value $Q=1$ corresponds to critical damping. 

The table below summarizes the value of the parameters. Though the values are reasonable, we do not expect them to correspond to any real device and corrected ones will come from a future validation process.
\begin{table}[htbp]
 \begin{tabular}{|c|c|c|}
  Parameter & Value & Units \\ \hline
	$c$ ($\nicefrac{\mathcal{C}}{m}$) & $2.919\times 10^{-9}$ & $N\cdot m^{4} \cdot kg^{-1}$\\
	$k$ ($\nicefrac{\mathcal{K}}{m}$) & $439.3$& $N\cdot m^{-1} \cdot kg^{-1}$ \\
	$d$ ($\nicefrac{\gamma}{m}$) & $[0, 20.96]$ & $N\cdot s \cdot m^{-1} \cdot kg^{-1}$\\
	$x_0$ & $1 \times 10^{-2}$ & $m$ \\
 \end{tabular}
\caption{The table summarizes the values of the parameters used for the numerical results reported in the text. Values for $d$ correspond to the range of damping ratio $Q \in [0,0.5]$.}
\label{tab:constants}	
\end{table}
\section{Results}
\label{sec:results}
\begin{figure}[htbp]
\begin{centering}
	\includegraphics[width=.6\textwidth]{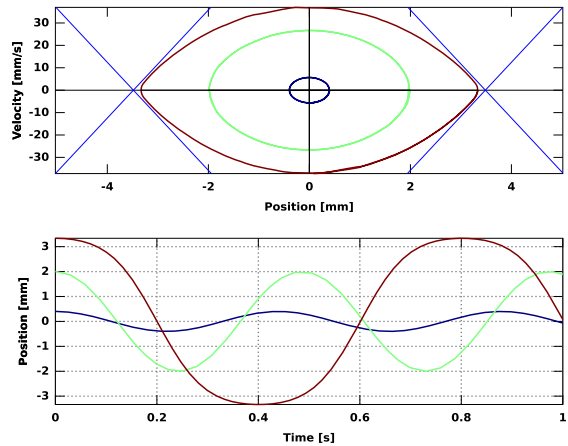}
	\caption{Trajectories in phase space and time series for the undamped system starting from three different initial condition. The frequency of the signal decreases with the amplitude due to the interaction of the magnets.}
	\label{fig:PhsSpace}
\end{centering}	
\end{figure}

\begin{figure}[htbp]
	\includegraphics[width=.8\textwidth]{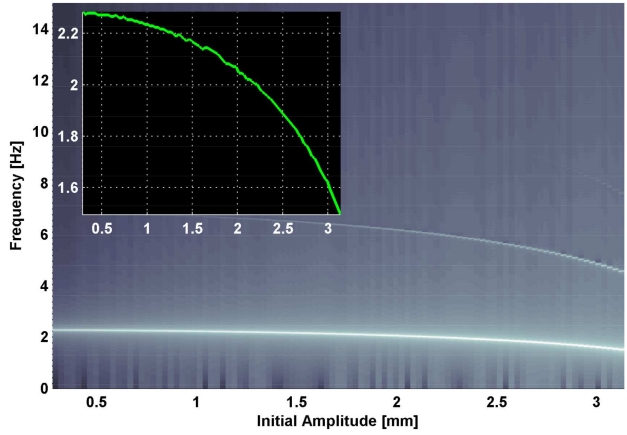}
	\caption{Variation of the natural frequency with the amplitude of the oscillations. The power spectrum of the orbits is plotted, the behavior of the main component is shown in detail in the inset.}
	\label{fig:NatFreq}
\end{figure}

\paragraph*{Phase space and time series.}
Numerical results for the undamped system are presented in Fig. \ref{fig:PhsSpace}. We take three initial conditions on the region of the phase space to study. All the initial conditions start with zero velocity, i.e. they lay on the horizontal axis. It is important to note that the time series of the fin displacement clearly show different frequencies. This is due to the attraction of the magnets, the higher the initial displacement the lower the frequency of the orbit. These results are shown in detail in Fig. \ref{fig:NatFreq}. For each initial condition we plot the power spectrum of the signal and it is visible how the main component decreases at higher amplitudes.

\begin{figure}[th!]
	\includegraphics[width=.6\textwidth]{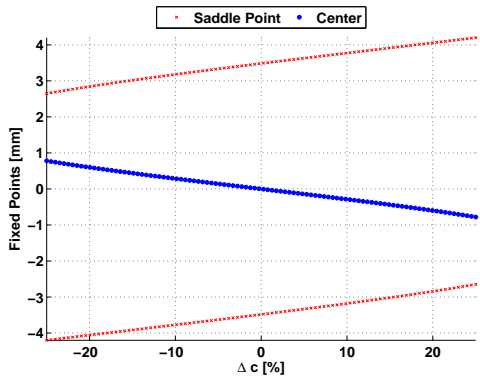}
	\caption{Control of the position of the center and saddles. The center moves symmetrically around the origin for differences $\Delta c$ between the magnetic constants. The maximum amplitude is also compromised, because the saddle on the side of the stronger magnet come closer to the center.}
	\label{fig:FPParam}
\end{figure}

The offset of the oscillations corresponds to the position of the center. By breaking the symmetry of the system, either by setting $c_1 \neq c_2$ or by feeding constant current to the solenoids, we can move the center off the origin. This could be required for turning maneuvers or useful for initiating oscillations. In Fig. \ref{fig:FPParam} we show how the center and the saddles move for different values of $c_1$ and $c_2$ (or increasing current).

\begin{figure}[h!t]
	\includegraphics[width=\textwidth]{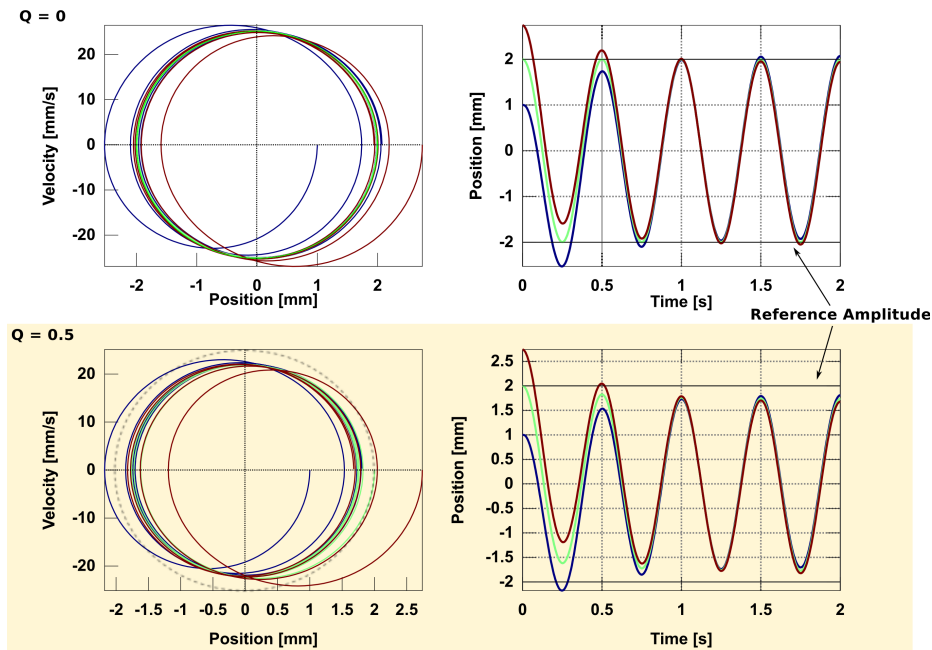}
	\caption{Trajectory in phase space of the actuated system for three different initial conditions close to the desired amplitude. On Top the results without damping and below them, with damping ratio of $Q=0.5$, which gives an unfeasible solenoid. In the second case the reference trajectory in phase space is shown in dashed line. On the left we show the time series of the displacement of the fin.}
	\label{fig:ActuatedPhsSpace}
\end{figure}

\paragraph*{Actuated system.}
The controller is provided with reference signal of the form $A\cos(wt+\phi)$. The amplitude $A$ is taken inside the bounds defined by the saddle-nodes. Given a value for the amplitude, we use the curve shown in the inset of Fig. \ref{fig:NatFreq} as a frequency lookup table. The initial phase $\phi$ is calculated from the initial conditions. We took the same initial conditions as in the previous section, namely $x(0)={A, 0.5\cdot A, 1.25 \cdot A}$. However simple, the control technique shows a remarkable performance as can be appreciated in Fig. \ref{fig:ActuatedPhsSpace}. In the figure the reference trajectory is the dashed line, for the case without damping, the actual trajectories of the system overlap with. On the left panels the time series of the displacement are shown and compared with the reference amplitude.

\paragraph*{Solenoid.}
If we take the maximum of the output from the controller, $F_c(t)$, and limit the maximum current fed to the solenoids to $I \leq 20 mA$, we can find a suitable expression for calculating the ideal number of turns $N$ of a coil. The net force generated on the fin magnet is given by the sum of forces due to the 2 coils, the wiring of which ensures that the force generated by each on the fin magnet is additive. This force can be expressed as, 
\begin{equation}
\begin{split}
 F_m = \frac{3}{2}\left(B_rR^2 \ell\right) \left(N I A\right) \operatorname{G}(x_m)\\
\operatorname{G}(x_m) = \frac{\left(x_m+x_0\right)^4 + \left(x_m-x_0\right)^4}{\left(x_m^2-x_0^2\right)^4},
\end{split}
\end{equation}
\noindent where $B_r$,$R$ and $\ell$ are as in \ref{eq:Magconstant}, and $x_m$ is the displacement corresponding to the maximum force $F_m$. Solving for $N$ and replacing with the corresponding values we get $N \in [600, 1000]$, for $Q \in [0,0.2]$. Higher values of damping impose too many turns on the solenoid. Although the model is not yet validated this result is encouraging.

\section{Discussion and conclusion}
\label{sec:conclusion_and_discussion}
In this short report we have presented the first step towards the design and construction of a novel actuator for small swimming machines. Though we have used rough models, the results show that we are pointing in the right direction. The design shown here is not necessarily the best in reducing the actuation needed. For example, one could think of using the instability of the center to initiate motion, by forcing the system through its bifurcation. This could be achieved by on-line modification of the distance between the supports or by bringing the permanent magnets closer. Additionally, placing a permanent magnet perpendicular to the plane at the origin could be used to further reduce the frequency of the orbits or to control the offset in a more sensitive way than the one shown here.

Our model includes dissipation proportional to the velocity and therefore the role of dissipation is marginal. More detailed models of the fluid dynamics and the bending of the fin will surely bring dissipation into a more primary role in the behavior of the system. In addition, thrust, heat dissipation and energy consumption could be estimated in such multi-physics models.

We have shown how a simple PID controller could perform reasonably when information about the phase portrait of the system, like the dependence of frequency with amplitude, is included. The use of adaptable frequency oscillators\cite{Righetti06b} or standard model-based controllers (like feedback linearization), could improve performance and reduce these requirements. Additionally, a system that is too flexible does not possess orbits without a controller. Such a controller would requires large amount of actuation, since it is \emph{forcing} the system to behave unnaturally. Therefore existence of orbits can be exploited to reduce energy consumption. This stresses the fact that passive dynamics are a key to improve the way we control and design our robots. Controlling the force between solenoids and moving magnets, brings several challenges on the design of the electrical circuits due to the changes in impedance. Another interesting aspect of the problem that will be addressed in further studies. We understand that results obtained solely from simulations are as 'words without actions', however the use of simple models can help us evaluate the feasibility of certain designs. In our particular case, a device with low friction could be driven with a tuned PID, a frequency lookup table and a solenoid with $~800$ turns, consuming about $20 mA$.

\section*{Acknowledgments}
We want to thank our lab coworkers Dr. Max Lungarella, Dr. Hugo G. Marques, Tao Li, Cristiano Alessandro and Dr. Lijin Aryananda for their comments and constructive criticism.
We thank Dr. Rolf Pfeifer for his continuous support to our research. Also for the continuation of the A.I. Lab., the friendly human environment which makes our work possible.
Funding for this work has been supplied by SNSF project no. 122279 (From locomotion to cognition) and by the European project no. FP7-231608 (OCTOPUS).
\paragraph*{Author Contributions} Both authors contributed equally to the work presented in this paper.

\bibliography{MagEReferences}

\end{document}